\documentclass{article}

\usepackage{arxiv}

\usepackage[utf8]{inputenc} 
\usepackage[T1]{fontenc}    
\usepackage{hyperref}       
\usepackage{url}            
\usepackage{booktabs}       
\usepackage{amsfonts}       
\usepackage{nicefrac}       
\usepackage{microtype}      
\usepackage{lipsum}
\usepackage{graphicx}
\graphicspath{ {./images/} }

\title{
Data-Driven Energy Modeling of Industrial IoT Systems: A Benchmarking Approach \thanks{This is a pre-print of a paper accepted for presentation at
the 30th IEEE Symposium on Computers and Communications (ISCC) 2025.}}

\author{
 Dimitris Kallis \\
  Department of Computer Science\\
  University of Cyprus \\
  \texttt{dkalli01@ucy.ac.cy} \\
   \And
 Moysis Symeonides \\
  Department of Computer Science\\
  University of Cyprus\\
  \texttt{msymeo03@ucy.ac.cy} \\
  \And
 Marios D. Dikaiakos \\
 Department of Computer Science\\
  University of Cyprus\\
  \texttt{mdd@ucy.ac.cy} \\
}

\begin{document}
\maketitle
\begin{abstract}
The widespread adoption of IoT has driven the development of cyber-physical systems (CPS) in industrial environments, leveraging Industrial IoTs (IIoTs) to automate manufacturing processes and enhance productivity. The transition to autonomous systems introduces significant operational costs, particularly in terms of energy consumption. Accurate modeling and prediction of IIoT energy requirements are critical, but traditional physics- and engineering-based approaches often fall short in addressing these challenges comprehensively.
In this paper, we propose a novel methodology for benchmarking and analyzing IIoT devices and applications to uncover insights into their power demands, energy consumption, and performance. To demonstrate this methodology, we develop a comprehensive framework and apply it to study an industrial CPS comprising an educational robotic arm, a conveyor belt, a smart camera, and a compute node. By creating micro-benchmarks and an end-to-end application within this framework, we create an extensive performance and power consumption dataset, which we use to train and analyze ML models for predicting energy usage from features of the application and the CPS system.
The proposed methodology and framework provide valuable insights into the energy dynamics of industrial CPS, offering practical implications for researchers and practitioners aiming to enhance the efficiency and sustainability of IIoT-driven automation.

\end{abstract}


\section{Introduction}

Exponential advances in Internet of Things (IoT) technologies have led to a rapid proliferation of IoT applications deployed in numerous domains, such as transportation, healthcare, industrial control, etc. Modern industrial infrastructures increasingly comprise cyber-physical systems (CPS), which integrate industrial devices with computational components and processes~\cite{Sisinni2018}.
Typically, CPS consist of IoT sensors, which monitor physical processes and systems to collect data reflecting their state, and of IoT actuators, which control physical systems to complete selected tasks. The management of CPS systems is performed by software components, which implement various computational processes like data analytics and Machine Learning~(ML) inference to provide CPS-component control and coordination. These software components are typically deployed in computing nodes within small data centers located near the IoT installations~\cite{Symeonides2022}. 

As CPS proliferate in distribution and scale, their operation is expected to take an increasing 
portion of the total energy consumption of IT systems, contributing significantly to global electricity consumption and $CO_2$ 
emissions. Consequently,  many recent research efforts focus on analysing and modeling the energy profile of edge computing devices 
and application workloads, and on adding energy-efficiency features to edge computing operations~\cite{pahlevan:tsc2021,greenscale2023,daSilva:tsc2023,dikaiakos:ccpe2023}. Most of these efforts, however,
focus on profiling computational workloads running on individual IoT computing devices.
Few studies focus on analyzing CPS installations that comprise both computing as well as sensing and actuating components working in tandem to perform complex tasks. Such scenarios are nevertheless prevalent in industrial infrastructures and  applications, particularly where Industrial IoT (IIoT) deployments automate manufacturing processes within industrial production pipelines. Such automation enables modern factories to operate around the clock, reducing personnel costs and enhancing productivity. With the prices of robotic arms dropping fast (46.2\% reduction between 2017 and 2021~\cite{aiindex:stanford:2024}), analysts predict that
such devices will become ubiquitous and could unleash a huge potential for the global economy~\cite{comingwave:2024}.
Given that power requirements significantly impact the operational costs of IIoT deployments, it is crucial to explore the power, energy, and performance of such applications, considering both their computational and physical components and performed tasks. However, studies of IIoT deployments 
often rely on simplistic models or focus on specific industrial machinery 
that do not consider real-world applications
~\cite{Vergnano2012, Paryanto2014, Xiaobin2022}. 
More comprehensive and realistic models are required to better predict 
energy consumption, 
improving the efficiency and cost-effectiveness of IIoT systems.

In this paper, we present a methodology and a benchmarking framework for analyzing the energy profile of CPS configurations in industrial automation systems, and for developing predictive models of their energy consumption. We deploy and demonstrate this framework to analyze the power and energy profile of an industrial CPS system, comprising a robotic arm, a conveyor belt, and a smart camera, collaboratively performing automated object sorting based on visual attributes such as color. 
The methodology is supported by our benchmarking and monitoring framework, which analyzes the energy profile of instructions submitted for execution by the application to the physical (e.g., robotic arm, belt) and digital (e.g., camera) components of the installation. 
The analysis maps component energy profiles, links them to operational parameters, and trains models to predict the energy profile of end-to-end applications based on features like speed, acceleration, and~weight.


The rest of the paper is structured as follows: Sec.~\ref{sec:relatedwork} introduces the related work and Sec.~\ref{sec:background} our cyber-physical setup. Next, Sec.~\ref{sec:methodology} and~\ref{sec:workloads} show our methodology and its workloads, respectively. In Sec.~\ref{sec:analysis} and~\ref{sec:modeling}, we apply exploratory analysis and the AI/ML training, while Sec.~\ref{sec:conclusion} concludes the paper.

\section{Related Work}
\label{sec:relatedwork}

The initial efforts to model the energy consumption of mechanical robotic components are related to the optimization of multi-robot systems~\cite{Vergnano2012}. 
However, the models and the analysis of these cases were relatively simple, taking into account only the initial and final trajectory points of a movement without considering the user-defined application. 
Recognizing this gap, Heredia et al~\cite{Juan2021} introduced an instruction-based approach to model the energy consumption of robots. By determining the energy consumed by each instruction, this approach enables the calculation of the overall energy consumption for any user-defined program. Building on this foundation, Heredia et al in~\cite{Juan2023}, proposed a novel optimization strategy tailored for lightweight robots, incorporating three energy-minimizing techniques: manufacturer command optimization, motion time determination optimization, and dissipative energy reduction. Additionally, in~\cite{Juan2023b}, they developed an energy consumption disaggregation pipeline, demonstrating its application on four robotic arms from various manufacturers.


Beyond mathematical formulation and modeling of robotic components' energy consumption, other approaches have emerged.  
For instance, the authors of \cite{Xiao2024} propose a methodology for developing models that map parameters such as joint torque and joint angular velocity of a robotic welding machine subsystem to its corresponding electric current and voltage. 
The authors in~\cite{Ke2018}, introduce a digital twin system to model the energy characteristics of robotic components, enabling the replication and analysis of results within a simulated environment.
Yan et al~\cite{Yan2021} and Ming et al~\cite{Ming2022}, adopt deep learning (DL) and transfer learning techniques to 
to model the energy profiles of robotic systems, with Yan et al. focusing on fine-tuning a neural network trained on one robotic arm to achieve high accuracy on a different system, while Ming et al. employ a ResNet-based model to evaluate the effectiveness of transfer learning in energy prediction tasks.

In summary, mathematical formulations cannot model complex robotic systems that involve multiple parameters and cyber-physical components.
Despite offering improved results in specific scenarios, approaches like digital twins, simulations, and DL, suffer from critical limitations: (i)~they do not offer any methodology generalization;~(ii)~they are tightly coupled with simulated environments; or (iii)~they utilize DL models, which offer high-accuracy, but can not provide explainability, making the optimization of energy consumption extremely difficult. In contrast, \textit{our methodology overcomes these limitations by offering a generalized, explainable, and adaptable solution that can be applied across various systems advancing the field of robotic energy consumption modeling.}


\section{Cyber-Physical Setup}
\label{sec:background}

\begin{figure}[t]
\centering
\includegraphics[width=0.8\textwidth]{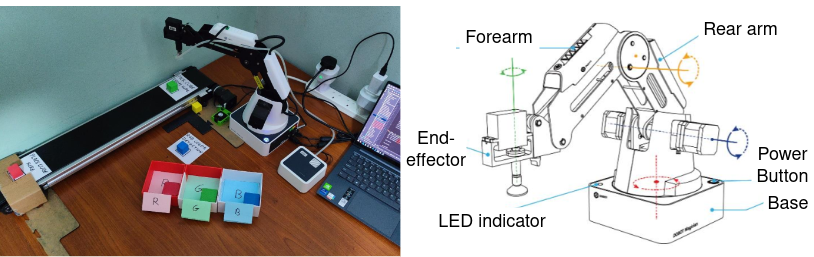}
    \vspace*{-.5\baselineskip}

\caption{Experimentation Setup \& Robotic Arm}
\label{fig:setup}
\vspace*{-1.3\baselineskip}
\end{figure}

Recognizing the gap in current state-of-the-art, we create a cyber-physical testbed~(Fig.~\ref{fig:setup}--left), which comprises both actuators of a robotic toolset, namely a robotic arm, suction cup end-effector, and a conveyor belt, and sensors, namely a smart camera and three smart plugs. Moreover, a laptop is used to coordinate execution and retrieve the monitoring data. 

\noindent \textit{\textbf{Dobot Magician Toolset:}}
Dobot Magician\footnote{\url{https://www.dobot.nu/en/product/dobot-magician-basic/}}, shown in Fig.~\ref{fig:setup}~(right), is a multipurpose educational robotic arm~\cite{dobot2020}. It is composed of a base, rear arm, forearm, and an end-effector, supports changeable end-effectors such as a suction cup and a gripper, and comes with 
kits like the conveyor belt~\footnote{\url{https://www.dobot.nu/en/product/dobot-conveyor-belt-kit/}}, which allows
the implementation of small-scale industrial production lines. 
As noted in Dobot's documentation~\cite{dobot2020}, the robotic arm achieves a maximum rotational speed of 320°/s for the rear arm, forearm, and base, while the servo motor can reach a rotational speed of 480°/s 
with a 250g payload.
Dobot software offers graphic programming methods, script programming, and more. Users control the robotic arm using either joint-based movements, defined by motion joints, or Cartesian movements, defined by X, Y, and Z coordinates.
Dobot's API lets users set velocity and acceleration as normalized values from 0 (no movement) to 100 (max speed or acceleration).
Finally,~the~suction~cup~actions~indicate~only~whether~it~is~enabled,~and~the~conveyor~belt~actions~enable~movement~and~set~the~speed~in~mm/s.

Our setup is depicted in Fig.~\ref{fig:setup} and is equipped with a suction cup end-effector (with air pump) and a conveyor belt. 
To use Dobot Magician, we leverage its API to control and automate the robotic arm, including its movements, conveyor belt operations, and end-effector actions. In our experiments we opted to use the joint coordinate system instructions, which direct the movement of the corresponding joints along with movement's velocity and acceleration. 

\noindent \textit{\textbf{Smart Camera:}} The selected robotic arm toolset does not provide any sensor for observing its environment. For that reason, we introduce a smart camera in our setup, specifically, a JeVois-A33~\footnote{\url{http://jevois.org/doc/Hardware.html}}, which includes a camera sensor, an embedded quad-core computer, and a USB video link in a tiny, self-contained package of 1.7 cubic inches weighing 17 grams. 
The camera can perform various ML vision tasks by loading algorithms via the JeVois Inventor program, processing each frame, and outputting the results. It supports popular open-source computer vision libraries, like OpenCV, TensorFlow, and Caffe. 
JeVois camera appears to the host computer as a conventional USB webcam and is entirely plug-and-play. 
The communication between JeVois and the application is done via serial messages, through which the uploaded algorithm (script) is executed in the JeVois camera, and the output shares information with the controller program. 
Examples of such scripts include object identity (color, type, classification), location (in 3D space), and detection count, among others.
For example, the color module used in our experiments is a script that captures camera frames, detects their colors, and sends the results to the control software. 

\noindent \textit{\textbf{Smart Plugs:} } In order to capture the energy consumption of the setup's subcomponents, we use four Meross smart plugs~\footnote{\url{https://www.meross.com/en-gc/smart-plug/alexa-smart-plug/3}}.
We selected this model because it exposes an API easily accessible over the network. So, we connect each actuator (Dobot robotic arm, belt, and suction compressor) to a smart plug.
It should be noted that the smart camera of our setup is powered through a USB connection. So, connecting the camera directly to a Meross smart plug is not possible. Thus, we used a USB hub, connected to a smart plug, to power the camera. 
As we described in Sec.~\ref{sec:smartcamera}, the JeVois camera has minimal and stable power consumption, with and without running our use-case ML algorithm, so, we exclude the camera's energy consumption from our results~and~analysis.

\section{CPS Benchmarking Methodology}
\label{sec:methodology}

\begin{figure}[t]
\vspace*{-.2\baselineskip}
\centering
\includegraphics[width=0.6\textwidth]{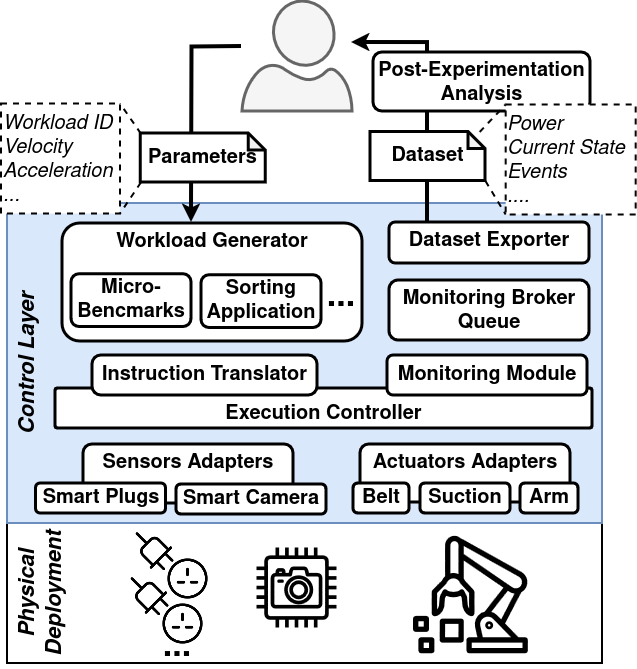}
    \vspace*{-.8\baselineskip}

\caption{High-level Overview of the Architecture} \label{fig:deployment}
\vspace*{-1.5\baselineskip}
\end{figure}

Fig.~\ref{fig:deployment} depicts our benchmarking and analysis methodology. 
To perform our trials, we build a software-based \textit{Control Layer}, between users and physical components, which abstracts the APIs of actuators and sensors allowing users to: (i)~repeatably submit various workloads with different configurations, (ii)~easily introduce new workloads and physical components, and (iii)~extract monitoring datasets of each run.

Our methodology starts with users submitting a set of experiment \textit{parameters} to the control layer. These \textit{parameters} include the selected workload and configuration preferences, which will be translated into low-level physical deployment configurations, such as the belt speed, the arm velocity, arm acceleration, etc. Having these preferences defined, the \textit{Workload Generator} loads the code of the selected application scenario and configures the system. 
Then, the \textit{Execution Controller} executes the workload on the physical infrastructure. To do that, it invokes the \textit{Instruction Translator} submodule, which translates a workload into instructions readable from physical components of the infrastructure. 
These instructions are executed by the \textit{Execution Controller} utilizing the \textit{Sensors \& Actuators Adapters}, which abstract the APIs of physical sensors~(e.g., camera) and actuators~(e.g., robotic arm). 

In our implementation, we built three \textit{Actuator Adapters} to handle the Dobot equipment, namely: (i)~the \textit{Arm Adapter}, which is responsible for coordinating the arm movement (joint movements, acceleration, and speed); (ii)~the \textit{Belt Adapter}, which controls the belt movement (e.g, on/off, and speed); and (iii)~the \textit{Suction Adapter}, which enables or disables the suction end-effector. 
These components use the Dobot Magician API in order to send the respective commands to the physical setup. 
Moreover, we implement two \textit{Sensor Adapters}: a \textit{Smart Camera} and a \textit{Smart Plug Adapter}. 
\textit{Smart Camera Adapter} allow us to deploy algorithms on the camera and retrieve their results at runtime. In our experiments, the color detection module is selected allowing us to extract the color of objects that are in front of the camera. 
The \textit{Smart Plug Adapter} retrieves the power metrics from the smart plugs using their API and each plug's identifier.  

At runtime, the \textit{Execution Controller} not only coordinates the physical execution, but also logs every command dispatched to the physical setup~(e.g., the belt is activated), the status of the physical infrastructure~(e.g., robotic arm positioning), and monitoring metrics~(e.g., power measurements from smart plugs). 
To this end, the \textit{Execution Controller} makes use of its \textit{Monitoring Module}, which extracts the latter metrics, introduces respective timestamps, and disseminates them to the \textit{Monitoring Broker Queue}. 
For this component, we utilize RabbitMQ\footnote{\url{https://www.rabbitmq.com/}} message queue and created a different topic for each metric. 
The asynchronous dissemination of the data to the broker guarantees the non-blocking execution of the commands on physical deployment. 
Then, the \textit{Data Exporter} module listens to all metric queues and whenever a new data point is received, creates a new row to a CSV file, which represents the dataset of a trial and includes physical deployment ``snapshots''~(every second), e.g., positioning of arm, suction status (true/false), belt status (true/false), camera detection (true/false), power measurements,~etc. 




\section{Cyber-Physical Workloads}
\label{sec:workloads}


\subsection{Micro-benchmarks}
\label{sec:microbench}
To evaluate separately the energy needs of each component, we built simple tasks dedicated to each physical component, namely: (i)~\textbf{Robotic Arm} that includes the movement of the robotic arm from a position \textit{A} to a position \textit{B} and back, with users being able to change the \textit{acceleration} and the \textit{speed} of this movement; (ii)~\textbf{Camera}, which considers the camera with the color detection algorithm running and includes two different states of the camera, namely, an object located in front of the camera and the \textit{color detected}, and \textit{absence} of any object and color; (iii)~\textbf{Belt} that can be started (movement), with its only parameter being the \textit{belt speed}; and (iv)~\textbf{Suction (End-effector)}, which enables or disables the suction of end-effector, so the value can be \textit{True} or \textit{False}. 
Finally, a physical parameter that we manually test, is the \textbf{payload} with different weights~(up to~730g) picked by the end-effector.

\subsection{End-to-end (Sorting) Application}
\label{sec:application}
We also evaluate a generic sorting application (convey-pick-classify-sort) scenario, which is common
in many industrial settings. In this generic sorting task, the conveyor belt moves objects, one by one in a predefined position. 
At this position, the arm grabs an object and puts it in front of a smart camera, which identifies its properties, and, based on this identification, the arm grasps  and places the object into a designated bucket.

To implement this scenario, we created an end-to-end sorting application pipeline. 
First, a colored cube is placed manually at the beginning of the conveyor belt (\textit{Step 1}). Then, the controller sends a command to the conveyor belt to transfer the cube close to the robotic arm (\textit{Step 2}). After that, the robotic arm picks the cube from the conveyor belt and places it in front of the smart camera to detect the color (\textit{Step 3}). Once the cube color is detected by the controller (\textit{Step 4}), a command is propagated to the robotic arm to pick the cube and place it into the appropriate box based on the color (\textit{Step 5}). The process is repeated until no cube is detected by the smart camera.
Since the execution is in loops, we define \textit{one application round} as the actions that occur between Step 1 and Step 5. 
Moreover, the configurations that one can change in the application execution are: (i)~\textbf{Arm Velocity}~(30-100); (ii)~\textbf{Arm Acceleration}~(20-100); and (iii)~\textbf{Belt Speed}~(10-80). 
The ranges are chosen to ensure trial functionality\footnote{For velocity under 30 and acceleration under 20, latency was unacceptable.}. 
Lastly, we also changed physical parameters, like \textbf{payload} weight.

Our system is modular and flexible, as we can easily implement new applications by defining the necessary instructions for their tasks and the parameters for each physical component involved. 
For example, switching from a sorting application to an object selection, a user would only require adjusting specific tasks~(e.g., item recognition camera module) and, potentially, the physical setup's organization (e.g., the arm's end-effector). 

\section{Exploratory Analysis}
\label{sec:analysis}
In this Section, we highlight the results of our exploratory analysis extracting useful insights for the submitted workloads. 

\begin{figure}[t]
\centering
\includegraphics[width=0.6\textwidth]{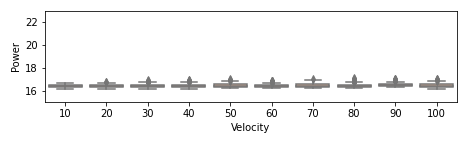}
    \vspace*{-1.3\baselineskip}

\caption{Power Demand (Watts) for Velocity Variations (in \%)} \label{fig:velocity}

\vspace*{-.2\baselineskip}
\centering
\includegraphics[width=0.6\textwidth]{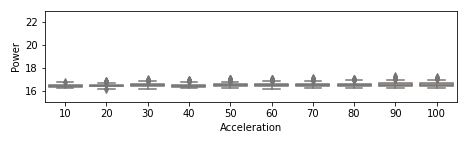}
    \vspace*{-1.3\baselineskip}

\caption{Power Demand (Watts) for Acceleration Variations (in \%)} \label{fig:acceleration}

\vspace*{-.2\baselineskip}
\centering
\includegraphics[width=0.6\textwidth]{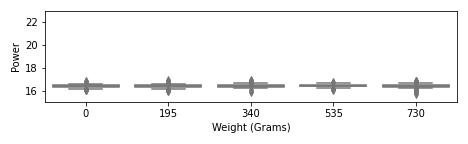}
    \vspace*{-1.3\baselineskip}

\caption{Power Demand (Watts) for Payload Variations (in kgrams)} \label{fig:payload}

\vspace*{-.3\baselineskip}
\centering
\includegraphics[width=0.6\textwidth]{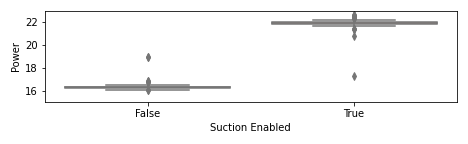}
    \vspace*{-1.3\baselineskip}

\caption{Power Demand (Watts) for Pump End-effector Status} \label{fig:pump}

\vspace*{-.2\baselineskip}
\centering
\includegraphics[width=0.6\textwidth]{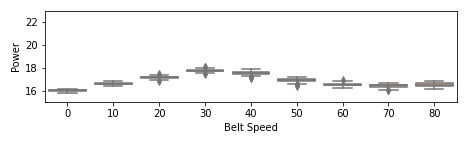}
    \vspace*{-1.3\baselineskip}

\caption{Power Demand (Watts) for Belt Speed Variations (in mm/s)} \label{fig:belt_speed}
\vspace*{-1.5\baselineskip}
\end{figure}

\subsection{Component-based Power Requirements}
Initially, we explore the power needs of each component (robotic arm, conveyor belt, and smart camera), by utilizing our micro-benchmarks~(see Sec.~\ref{sec:microbench}). 
So, we systematically change components' configurations and measure their power needs. 
By identifying power-hungry components of a CPS, designers of CPS apps can optimize their power consumption during the design phase. This approach eliminates the need to deploy a full application in the field to assess power requirements, enabling efficient and proactive power management.


\subsubsection{Smart Camera Color Detection} 
\label{sec:smartcamera}
To test the power needs of this module, we manually put a color cube in front of the camera and captured the power consumption, with and without the cube. 
The difference in the power consumption between detecting and not detecting color is minimal~($\approx$2.13~Watts).

\subsubsection{Robotic Arm Movement Velocity, Acceleration, and Payload}
Afterwards, we test the velocity and the acceleration of the robotic arm. As noted in Sec.~\ref{sec:background}, the API of the robotic arm allows us to set its velocity and acceleration in a normalized range 0-100. 
In this analysis, we change these values in increments of 10 until reaching the maximum. Fig.~\ref{fig:velocity} and \ref{fig:acceleration} show the box plots of power consumption measurements for different velocity and acceleration. Interestingly, medians ($\approx$16.5 Watts) and distributions of power consumption for these experiments seem not to  be influenced by velocity and acceleration levels. Similar results appear when we change the weights of the caring payload (Fig.~\ref{fig:payload}). 
Even if the Dobot manual determines the maximum weight being 500 grams~\cite{dobot2020}, we tested with payloads of up to 730 grams
without any problem (0, 195, 340, 535, and 730 grams).

\subsubsection{End-effector (suction cup)} has two stages, namely suction enabled equal to \textit{True} or \textit{False}. The results of our measurements are depicted in Fig.~\ref{fig:pump}. Interestingly, the suction effector, due to its pump, consumes the largest portion of power among the components, namely $\approx$22 watts when enabled. When the pump is turned off, the system's power consumption is reduced to around 16 watts, which is the idle power of the robotic arm.

\subsubsection{Belt Speed}
Finally, we evaluate the results when only the belt is working at different speed levels from 0 to 80 millimeters per second (mm/s). 
Fig.~\ref{fig:belt_speed} highlights the results of power measurements for different speeds. 
The power consumption of the system is increasing up to a 30-40~mm/s belt speed. 
After that point, the power decreases until the speed of the belt reaches 60~mm/s. Then, the power requirements reach a low level
and remain steady for the rest of the speed levels. 
This relation between speed and power may show the effect of 
factors such as motor efficiency curves, frictional dynamics, and mechanical resonance.

\subsubsection{Key Takeaways} 
The evaluation of different components and parameters on deployment's power needs indicates that: ~\textit{
(i)~the smart camera has stable power consumption when performing color inference; (ii)~arm velocity, arm acceleration, and payload parameters have almost zero influence on the power consumption; 
(iii)~the suction end-effector is the component with the highest power consumption when enabled (performing suction); and (iv)~the belt speed influences the power consumption in an unexpected way, most probably due to physical interaction between the belt's components.}

\subsection{Evaluation of the End-to-End Application}

In this part, we focus on the end-to-end (sorting) application as it is described in Sec.~\ref{sec:application}. 
Specifically, we executed the application workload with different input configurations, like arm velocity, arm acceleration, and belt speed. 
When we change the values of one parameter, the rest remain constant. 

Having performed a large number of trials and capturing the power consumption for every second of operation allows us to create a dataset with over 22k data points.
With this dataset in hand, we first evaluate how the configurations of different subcomponents impact the overall power requirements during end-to-end application execution. 
Then, we explore how variations in parameters affect both power and energy consumption ($energy = power \times duration$), providing insights into how different configurations can optimize energy efficiency.

 
\begin{figure}[t]
\centering
\includegraphics[width=0.6\textwidth]{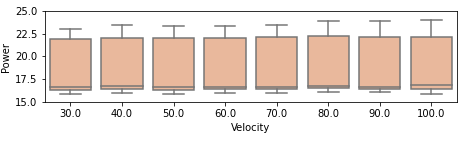}

    \vspace*{-1.3\baselineskip}
\caption{Application Power Demand (Watts) for Velocity Levels~(\%)} \label{fig:appvelocity}

\centering
\includegraphics[width=0.6\textwidth]{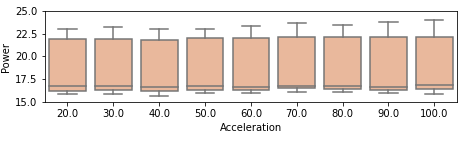}

    \vspace*{-1.3\baselineskip}
\caption{Application Power Demand (Watts) for Acceleration Levels~(\%)} \label{fig:appacceleration}

\centering
\includegraphics[width=0.6\textwidth]{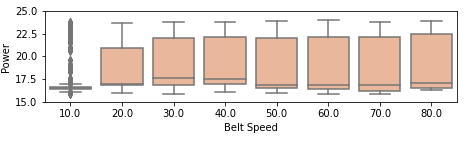}

    \vspace*{-1.3\baselineskip}
    \caption{Application Power Demand (Watts) for Belt Speed~(mm/s)} \label{fig:appbelt}


    \vspace*{-1.3\baselineskip}

\end{figure}

\subsubsection{Component contribution in Application Power needs:} Fig.~\ref{fig:appvelocity} and~\ref{fig:appacceleration} show that the changing arm velocity and acceleration do not alter the distribution of the power with median power to be slightly less than 17.5 Watts, and the large majority of the values being between 16 and 22.5 Watts, in all cases.
When we change the belt speed, keeping a static velocity and acceleration of the robotic arm, we observe the median power consumption in Fig.~\ref{fig:appbelt} to follow a similar trend, like the values of Fig.~\ref{fig:belt_speed}. 
However, this does not significantly affect the range and distribution of power consumption values during the trials, with the exception of cases where the belt speed is set to 10 mm/s and 20 mm/s. At a belt speed of 10 mm/s, we observe a large number of outliers, and at 20 mm/s, the upper quantile is slightly lower compared to the other results. 
Intuitively, this may indicate that at lower belt speeds, the rest of the system is relatively idle. 
When the cube reaches the designated point, and the robotic arm and the end-effector activate, there is a noticeable increase in power consumption, resulting in these outliers. 
In the remaining trials, such anomalies are not observed because low-power consumption points (when only the belt is moving) are less frequent.
Thus, the power needs of specific components may be absorbed during the execution of a CPS~app.




\begin{figure}[t]
\centering
\includegraphics[width=0.6\textwidth]{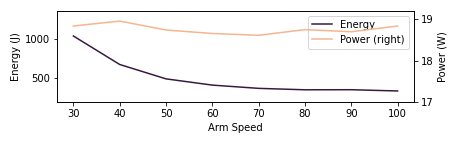}
    \vspace*{-1.3\baselineskip}

\caption{Energy (Joules) \& Power Demand (Watts) for Velocity Levels} \label{fig:energspeed}

\centering
\includegraphics[width=0.6\textwidth]{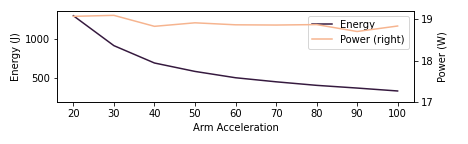}
    \vspace*{-1.3\baselineskip}

\caption{Energy (Joules) \& Power Demand (Watts) for Acceleration Levels} \label{fig:energyacc}

\centering
\includegraphics[width=0.6\textwidth]{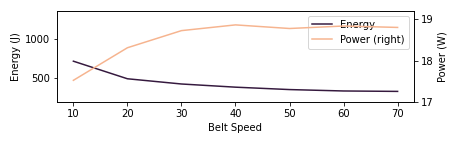}
    \vspace*{-1.3\baselineskip}

\caption{Energy (Joules) \& Power Demand (Watts) for Belt Speed (mm/s)} \label{fig:energybelt}
    \vspace*{-1.\baselineskip}

\end{figure}

\begin{figure}[!t]
\centering
    \begin{minipage}{.33\textwidth}
         \includegraphics[width=1\textwidth]{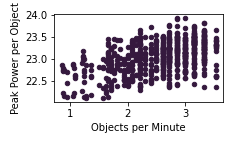}
            \vspace*{-1.9\baselineskip}
        
    \end{minipage}%
    \begin{minipage}{.33\textwidth}
\includegraphics[width=1\textwidth]{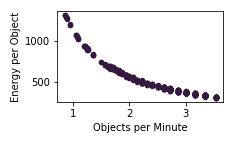}          
    \vspace*{-1.5\baselineskip}

    \end{minipage}
        \caption{Peak Power (Watts) and Energy (Joules) per object vs Throughput}
      \vspace*{-1.5\baselineskip}
\label{fig:throughput}
\end{figure}

\subsubsection{Energy and Power needs for different configurations} Here, we compute the energy consumption per application round for different parameterizations (arm velocity, arm acceleration, and belt speed). 
The energy of a round is computed by multiplying the mean power usage of the system with the duration of the round measured in seconds.
Fig.~\ref{fig:energspeed},~\ref{fig:energyacc}, and~\ref{fig:energybelt} highlight the results for the arm speed, arm acceleration, and belt speed, respectively. 
The primary y-axis depicts the computed energy in Joules and the secondary y-axis depicts the mean power consumption during the round in Watts.
As we can see, the power consumption of the overall deployment is relatively stable independently from the configuration of Arm Velocity and Acceleration. 
For the Belt Speed, the power consumption increases for speeds up to 40mm/s and, after that, the power becomes stable at about 19Watts.  
Interestingly, the energy is dominated by the speed and acceleration of physical components, and its drop follows a logarithmic trend. 

\subsubsection{Energy \& Peak Power per Object vs Throughput} 
The plots in Fig.~\ref{fig:throughput} illustrate the relationship between the application throughput for different input configurations, measured in objects per minute, and two key performance metrics: peak power consumption (Watts) per object and energy consumption (Joules) per object. 
The left plot shows 
that peak power is slowly increasing within a narrow range of approximately 22.5 to 24 watts, following the variations in the objects' processing speed. This suggests that the system's peak power demands are not significantly influenced by the application throughput. 
Moreover, the right plot demonstrates a clear inverse relationship between processed objects per minute and energy consumption per object.
Specifically, configurations that increase the number of objects processed per minute lead to a reduction in energy consumption per object. 
This trend was illustrated also in previous experiments, and highlights the energy efficiency gains achieved at higher processing speeds.

\subsubsection{Key Takeaways} 
Analyzing our datasets, we found that: ~\textit{(i)~changes in end-to-end application
parameters, such as belt speed, may slightly affect the median power requirement without altering the overall power demand distribution; 
and (ii)~the mean and peak power requirement for a complete cycle of the end-to-end application remains stable despite variations in speed of sub-components, but per-round energy consumption decreases logarithmically with belt speed.}

\section{Energy and Performance Modeling}
\label{sec:modeling}

In this section, we evaluate the potential use of ML/AI models to accurately estimate power requirements, predict energy consumption and latency based on end-to-end application parameters, and extract insights from feature importance analysis. Using a combined dataset of 22,384 rows representing the physical deployment state through 16 features, we trained 20 regression models, including linear regression, tree-based models, and ensembles, with the support of the PyCaret 
library. PyCaret optimized the models via hyperparameter tuning and provided metrics such as Mean Absolute Error (MAE), Mean Squared Error (MSE), Root Mean Squared Error (RMSE), Root Mean Squared Logarithmic Error (RMSLE), and Mean Absolute Percentage Error (MAPE), while also facilitating post-training feature importance analysis to explain the results.
Moreover, k-fold cross-validation with stratified sampling was employed, ensuring balanced representation of configurations across training and testing sets.

\begin{table}[t]
\vspace{0.07in}
\centering
\begin{tabular}{|p{3cm}|p{1cm}|p{1cm}|p{1cm}|p{1cm}|p{1.4cm}|p{1cm}|}
\hline
\textbf{Model} & \textbf{MAE} & \textbf{MSE} & \textbf{RMSE} & \textbf{R2} & \textbf{RMSLE} & \textbf{MAPE} \\ \hline
\multicolumn{7}{|c|}{\textbf{Models for Power State}} \\ \hline
Random Forest  & \textbf{0.69} & \textbf{1.66} & \textbf{1.29} & \textbf{0.78} & \textbf{0.06} & \textbf{0.036} \\ \hline
Extra Trees  & 0.70 & 1.88 & 1.37 & 0.75 & 0.06 & 0.036 \\ \hline
XGBoost & 0.83 & 1.91 & 1.38 & 0.75 & 0.06 & 0.044 \\ \hline
LightGBM & 0.88 & 1.99 & 1.41 & 0.74 & 0.06 & 0.046 \\ \hline
Decision Tree & 0.74 & 2.60 & 1.61 & 0.66 & 0.07 & 0.038 \\ \hline
\multicolumn{7}{|c|}{\textbf{Models for Round Energy}} \\ \hline
Random Forest & \textbf{20.14} & \textbf{744.99} & \textbf{27.22} & \textbf{0.9722} & \textbf{0.0537} & \textbf{0.0423} \\ \hline
XGBoost & 20.21 & 751.06 & 27.32 & 0.9720 & 0.0537 & \textbf{0.0423} \\ \hline
Extra Trees & 20.23 & 751.42 & 27.33 & 0.9720 & 0.0537 & 0.0424 \\ \hline
Decision Tree & 20.21 & 751.11 & 27.32 & 0.9720 & 0.0537 & 0.0424 \\ \hline
Gradient Boost & 20.69 & 760.38 & 27.51 & 0.9717 & 0.0544 & 0.0431 \\ \hline
\multicolumn{7}{|c|}{\textbf{Models for Round Duration}} \\ \hline
Random Forest & \textbf{1.0699} & \textbf{2.1431} & \textbf{1.4598} & \textbf{0.9712} & \textbf{0.0524} & \textbf{0.0423} \\ \hline
Extra Trees & 1.0714 & 2.1644 & 1.4667 & 0.9710 & \textbf{0.0524} & 0.0422 \\ \hline
Decision Tree & 1.0709 & 2.1643 & 1.4666 & 0.9710 & \textbf{0.0524} & 0.0422 \\ \hline
XGBoost & 1.0728 & 2.1661 & 1.4672 & 0.9709 & \textbf{0.0524} & 0.0423 \\ \hline
Gradient Boost & 1.0904 & 2.1872 & 1.4755 & 0.9707 & 0.0530 & 0.0431 \\ \hline
\end{tabular}
\caption{Performance Metrics for Various Models Across Three Tasks}
\label{table:combined}
\vspace*{-1\baselineskip}
\end{table}

\subsection{Power Requirements Modeling}

In this part, we train AI/ML models that take as input the system's state, e.g., positioning of the arm, its acceleration, its velocity, statuses of the belt and the suction (enabled or not), and predict the system's instantaneous power demand. 

\subsubsection{Modeling Performance} 
Table~\ref{table:combined} illustrates the top-5 models based on their $R^2$ and their performance.
The best model is the Random Forest providing a MAPE of approximately 3.6\%, followed by Extra Trees regressor and Extreme Gradient Boosting (XGBoost) with their MAPEs being 3.6 and 4.5\%, respectively. Even if Extreme Trees have the same MAPE as the Random Forest model, the latter has a higher $R^2$ value, which indicates a better fitting on the predicted data distribution. Finally, Light Gradient Boosting (LightGBM) and Decision Tree regressors offer lower performance with 0.74 and 0.66 $R^2$ scores, respectively. 

\subsubsection{Key Takeaways} \textit{ Out of the various models explored with our experimentation datasets, the tree-based regressors demonstrate better accuracy on this task, with the Random Forest regressor providing the best results with a 3.6\% MAPE.}

\subsection{Energy and Duration Estimators} 

Having the end-to-end application working on rounds~(see Sec.~\ref{sec:application}), next we build models that take as input round's parameters (velocity, acceleration, belt speed, and cube weight), and predict the energy consumption and the round duration. 





        


\subsubsection{Modeling Performance} 
Table~\ref{table:combined} also illustrate the top 5 models for energy consumption and round duration. 
In both cases, random forest is the best model with its MAPE being 4.23\% and $R^2$ being 97\%. The results are too similar because, at the computation of the energy, we multiply the duration of the round by the average power needs of the system. Since the average power needs of the system are relatively static (compared to the duration), the dominant parameter seems to be the duration of the workload.
Another observation is that all of the top-5 models utilize Decision Tree alterations in order to predict their results. 
In the case of the energy of each round, Extreme Gradient Boosting provides the same MAPE as Random Forest, but the rest of the metrics are slightly lower (e.g., $R^2$ is 0.972 and MAE is 20.21). Extra trees, Decision Tree, and Gradient Boosting follows with slightly lower MAPE (4.24\%, 4.24\%, 4.31\%). Generally, all models provide good results and their differences are negligible. 
For the round duration predictors, the Extra Trees model is in the second place with a MAPE equal to 4.22\% and $R^2$ equal to 97.10\%. Then, we have the Decision Tree, Extreme Gradient Boosting, and Gradient Boosting models to provide marginally lower results (e.g., MAPE 4.22\%, 4.23\%, 4.31\%). 

\subsubsection{Key Takeaways} Our methodology generated \textit{accurate models for round energy and duration, with the best of them (Random Forest) having an error up to 4.23\% in both target metrics, with other models providing comparable~results.}

\begin{figure}[t]
\vspace*{-.5\baselineskip}
\centering
\includegraphics[width=0.67\textwidth]{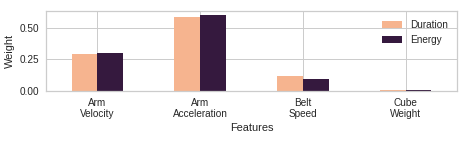}
    \vspace*{-1.4\baselineskip}

\caption{Feature Importance for Duration Model \& Energy Model} \label{fig:modelenergy}


    \vspace*{-1.3\baselineskip}
\end{figure}

\subsection{Feature Importance and Explainability} 
Next, we extracted from the best model of each case its feature importance and we depict them in Fig.~\ref{fig:modelenergy}.
Interestingly, the most important feature in every case is the arm acceleration, with its importance being between 0.60 and 0.58, for energy and duration models, respectively. 
In both cases, acceleration is followed by the arm velocity with 0.29 importance, and the third feature is the belt speed, with its importance being 0.08 and 0.11 for energy and duration models, respectively.
Finally, the cube weight is the least important feature, with almost no contribution.
The dominance of arm acceleration and velocity in feature importance likely stems from their impact on task timing rather than direct power requirements. In our setup, acceleration and velocity define the duration of specific actions, indirectly influencing the overall energy consumption of an application cycle ($energy = power \times time$). The relatively constant power demands of other components amplify this effect, making acceleration appear as a key determinant of energy consumption. 

\subsubsection{Key Takeaways} \textit{The most important feature for both energy and round duration is the robotic arm acceleration followed by arm velocity. Belt speed slightly influences the models and cube weight does not have any contribution.}

\section{Conclusion}
\label{sec:conclusion}

In this paper we introduce and implement a methodology to benchmark and analyze CPS applications, focusing on power needs, energy consumption, and performance. Using component- and application-based workloads on a robotic-arm/conveyor belt/smart camera-based system, we collected and analyzed an extensive dataset, revealing key insights: parameters like arm velocity, acceleration, and payload have minimal impact on power consumption, while the end-effector (pump) is the most power-intensive component.  Overall power consumption was stable, but power needs varied across different application operations and energy per application round decreased logarithmically with higher speed and acceleration.  
For AI/ML prediction modeling, our best models accurately estimated power, energy, and duration, achieving MAPE errors of $\approx$3.6\% for power prediction and 4.23\% for energy and duration prediction.  
Beyond accurate predictions, our ML approach helps optimize IIoT setups by identifying power-hungry components like the suction end-effector, enabling targeted cost reductions and energy-efficient adjustments.

Our future work includes validating our approach in larger, real-world industrial setups and integrating domain knowledge to enhance prediction accuracy. We also plan to explore advanced ML methods and benchmark against other approaches.

\footnotesize{\noindent \textbf{Acknowledgement.} This work is supported in part by EU's Horizon Europe programme through the AI-DAPT project (Grant No. 101135826) and the AdaptoFlow project via the TrialsNet Open Call (Grant No. 101017141).}

\bibliographystyle{IEEEtran}

\bibliography{conference_101719} 

\end{document}